\documentclass{article}
\usepackage{ijcai24}

\usepackage{amsmath,amsfonts,bm}

\def\eqref#1{equation~\ref{#1}}

\def\1{\bm{1}}

\def\ra{{\textnormal{a}}}

\def\rs{{\textnormal{s}}}

\def\rz{{\textnormal{z}}}

\def\vp{{\bm{p}}}

\def\evp{{p}}

\DeclareMathAlphabet{\mathsfit}{\encodingdefault}{\sfdefault}{m}{sl}
\SetMathAlphabet{\mathsfit}{bold}{\encodingdefault}{\sfdefault}{bx}{n}

\def\gA{{\mathcal{A}}}
\def\gB{{\mathcal{B}}}

\def\gK{{\mathcal{K}}}
\def\gL{{\mathcal{L}}}

\def\gS{{\mathcal{S}}}

\def\gU{{\mathcal{U}}}

\def\sG{{\mathbb{G}}}

\newcommand{\E}{\mathbb{E}}

\newcommand{\R}{\mathbb{R}}

\newcommand{\KL}{D_{\mathrm{KL}}}

\usepackage{times}
\usepackage{soul}
\usepackage{url}
\usepackage[hidelinks]{hyperref}
\usepackage[utf8]{inputenc}
\usepackage[small]{caption}
\usepackage{algorithm}
\usepackage{algorithmic}
\usepackage[switch]{lineno}

\renewcommand*{\thefootnote}{\fnsymbol{footnote}}
\author{
Xiang Zheng$^1$
\and
Xingjun Ma$^2$\and
Chao Shen$^3$\And
Cong Wang$^1$\footnote{Corresponding author.}\\
\affiliations
$^1$City University of Hong Kong\\
$^2$Fudan University\\
$^3$Xi'an Jiaotong University\\
\emails
\{xzheng235-c@my, congwang@\}cityu.edu.hk,
xingjunma@fudan.edu.cn,
chaoshen@mail.xjtu.edu.cn
}

\usepackage{graphicx,subcaption}
\graphicspath{{./figs/}}
\usepackage{booktabs,array,multirow,makecell}
\usepackage{amsmath,amssymb,amsfonts,bm,nicefrac}
\usepackage{cleveref}
\usepackage{amsthm}

\newcommand{\p}{$\,\pm\,$}
\usepackage{bm}

\usepackage{xcolor}
\definecolor{red}{RGB}{255,0,0}
\definecolor{green}{RGB}{2,128,0}
\definecolor{blue}{RGB}{11,0,255}
\newcolumntype{C}{>{\raggedleft\arraybackslash}p{0.12\textwidth}}

\title{Constrained Intrinsic Motivation for Reinforcement Learning}

\begin{document}

\maketitle
\renewcommand*{\thefootnote}{\arabic{footnote}}
\setcounter{footnote}{0}

\begin{abstract}
This paper investigates two fundamental problems that arise when utilizing Intrinsic Motivation (IM) for reinforcement learning in Reward-Free Pre-Training (RFPT) tasks and Exploration with Intrinsic Motivation (EIM) tasks: 1) how to design an effective intrinsic objective in RFPT tasks, and 2) how to reduce the bias introduced by the intrinsic objective in EIM tasks.
Existing IM methods suffer from static skills, limited state coverage, sample inefficiency in RFPT tasks, and suboptimality in EIM tasks.
To tackle these problems, we propose \emph{Constrained Intrinsic Motivation (CIM)} for RFPT and EIM tasks, respectively: 1) CIM for RFPT maximizes the lower bound of the conditional state entropy subject to an alignment constraint on the state encoder network for efficient dynamic and diverse skill discovery and state coverage maximization; 2) CIM for EIM leverages constrained policy optimization to adaptively adjust the coefficient of the intrinsic objective to mitigate the distraction from the intrinsic objective.
In various MuJoCo robotics environments, we empirically show that CIM for RFPT greatly surpasses fifteen IM methods for unsupervised skill discovery in terms of skill diversity, state coverage, and fine-tuning performance. Additionally, we showcase the effectiveness of CIM for EIM in redeeming intrinsic rewards when task rewards are exposed from the beginning. Our code is available at \url{https://github.com/x-zheng16/CIM}.
\end{abstract}

\section{Introduction}
\label{sec: intro}

In the realm of Reinforcement Learning (RL), Intrinsic Motivation (IM) plays a vital role in the design of exploration strategies in both Reward-Free Pre-Training (RFPT) tasks and Exploration with Intrinsic Motivation (EIM) tasks~\cite{barto2013intrinsic}.
It allows the RL agent to efficiently visit novel states by assigning higher intrinsic bonuses to unfamiliar states~\cite{zhang2021made}.
Current IM methods can be classified into three categories: knowledge-based, data-based, and competence-based IM methods~\cite{laskin2021urlb}. 

Knowledge-based and data-based IM methods are employed in both RFPT and EIM tasks to encourage the agent to explore novel regions.  Knowledge-based IM methods maximize the deviation of the agent's latest state visitation from the policy cover (i.e., the regions covered by all prior policies)~\cite{zhang2021made}. These methods commonly estimate the density of the policy cover via the pseudo-count of state visit frequency~\cite{bellemare2016unifying,fu2017ex2}, prediction errors of a neural network~\cite{pathak2017curiosity,burda2018exploration}, or variances of outputs of an ensemble of neural networks~\cite{pathak2019self,lee2021sunrise,bai2021principled}. Data-based IM methods, on the other hand, directly maximize the state coverage (i.e., the region visited by the latest policy) via maximizing the state entropy~\cite{hazan2019provably,mutti2021task,liu2021behavior,seo2021state}.
However, knowledge-based and data-based IM methods are inefficient in RFPT tasks since they do not condition the latent skill variable, limiting the fine-tuning performance of the pre-trained policy in downstream tasks~\cite{liu2021aps}. Moreover, when utilized in EIM tasks, these IM methods introduce non-negligible biases to the policy optimization, leading to suboptimal policies~\cite{chen2022redeeming}. Specifically, intrinsic objectives may result in excessive exploration even when the task rewards are already accessible. This distraction induced by intrinsic objectives can deteriorate the performance of the RL agent and impede the wider application of these methods in EIM tasks.

Competence-based IM methods are designed for unsupervised skill discovery in RFPT tasks. They primarily maximize the mutual information between the state representation and the latent skill variable to learn a latent-conditioned poilcy~\cite{gregor2016variational,sharma2019dynamics,laskin2022cic}. The policy conditioned on the latent skill variable is required to change the state of the environment in a consistent and meaningful way, e.g., walking, flipping, pushing, to be finetuned efficiently in the downstream tasks.
However, current competence-based IM methods have shown poor performance in the Unsupervised Reinforcement Learning Benchmark (URLB)~\cite{laskin2021urlb}, a benchmark of IM methods evaluated in RFPT tasks. Intuitively, directly maximizing the mutual information does not guarantee extensive state coverage or the discovery of dynamic skills and easily converges to simple and static skills due to the invariance of the mutual information to scaling and invertible transformation of the input variables~\cite{park2021lipschitz,park2023controllability}. Here, ``dynamic" skills refer to skills that facilitate large state variations, e.g., running for locomotion tasks and moving for manipulation tasks.
To address this limitation, Park et al.~\shortcite{park2021lipschitz} proposed Lipschitz-constrained Skill Discovery (LSD) to encourage dynamic skills. However, LSD suffers from severe sample inefficiency. The primary reason is that maximizing the intrinsic objective of LSD cannot guarantee maximum state entropy.

To overcome the limitations of existing knowledge-based, data-based, and competence-based IM methods, in this paper, we propose \emph{Constrained Intrinsic Motivation (CIM)} which is 1) a novel constrained intrinsic objective in RFPT tasks, i.e., a lower bound of the conditional state entropy subject to an alignment constraint on the state encoder network, to make the RL agent discover dynamic and diverse (distinguishable) skills more efficiently; 2) a Lagrangian-based adaptive coefficient for the intrinsic objective in EIM tasks to alleviate the performance decrease due to the bias introduced by the intrinsic rewards.

In summary, we make the following main contributions:
\begin{itemize}
	\item We propose \emph{Constrained Intrinsic Motivation (CIM)} to overcome the limitations of knowledge/data-based and competence-based IM methods by combining the best of both worlds. CIM outperforms state-of-the-art IM methods, improving performance and sample efficiency in multiple MuJoCo robotics environments.
	
	\item CIM for RFPT introduces a lower bound for the state entropy, conditioning the state entropy on the latent skill variable without compromising the power of maximum state entropy exploration. CIM for RFPT also introduces a novel alignment constraint on the state encoder network. Compared with LSD, our CIM reduces the number of required samples by 20x less (e.g., from 400M to 20M in the environment Ant). Besides skill diversity and state coverage, our CIM achieves the highest fine-tuning performance in the Walker domain of URLB.
	
	\item CIM for EIM derives an adaptive coefficient of the intrinsic objective leveraging the constrained policy optimization method. We empirically demonstrate that the adaptive coefficient can effectively diminish the bias introduced by intrinsic bonuses in various MuJoCo tasks and improve the average task rewards.
\end{itemize}

\section{Preliminaries}
\label{sec: pre}

\subsection{Markov Decision Processes} The discounted Markov Decision Process (MDP) is defined as $M=(\gS, \gA, P, R, \gamma, \mu)$, where $\gS$ and $\gA$ stand for the state space and the action space separately, $P: \gS \times \gA \to \Delta(\gS)$ is the transition function mapping the state $s$ and the action $a$ to the distribution $P(s'|s,a)$ in the space of probability distribution $\Delta(\gS)$ over $S$, $R: \gS \times \gA \times \gS \to \R$ is the reward function, $\gamma \in [0,1)$ is the discount factor, and $\mu \in \Delta(\gS)$ is the initial state distribution. We focus on the episodic setting where the environment is reset once the agent reaches a final state $s_f$, a terminated state within the goal subsets $\sG$ or a truncated state $s_T$. At the beginning of each episode, the agent samples a random initial state $s_0 \sim \mu$; at each time $t=0,1,2,..., T-1$, it takes an action $a_t \in \gA$ computed by a stochastic policy $\pi:\gS \to \Delta(\gA)$ or a deterministic one $\pi:\gS \to \gA$ according to the current state $s_t$ and steps into the next state $s_{t+1} \sim P(\cdot|s_t, a_t)$ with an instant reward signal $r = R(s_t, a_t, s_{t+1})$ obtained.

\subsection{Reward-Free Pre-Training and Exploration} RFPT and EIM are two types of intrinsically motivated RL tasks. To present the optimization objectives of RFPT and EIM, we first define the state distribution induced by the policy $\pi$ as $d_\pi(s)= (1-\gamma) \sum_{t=0}^{\infty} \gamma^t P(s_t=s|\mu, \pi)\in\gK$, where $\gK$ is the collection of all induced distributions. The extrinsic objective (the expectation of the task reward) is then $J_\text{E}(d_\pi)= \E_{s\sim d_{\pi}} \left[ r_\text{E} \right]$, where  $r_\text{E}=R_\text{E}(s, a, s')$ is the extrinsic task reward function. The intrinsic objective $J_\text{I}: \gK\to \R$ is defined as a differentiable function of the induced state distribution $d_\pi$ with $L-$Lipschitz gradients.

In RFPT tasks, the task reward $r^e$ is not available, and the agent aims to maximize only the intrinsic objective
\begin{equation}
	\label{eqn: RFPT}
	L_{k}^{\text{RFPT}}(\pi) = J_\text{I}(d_\pi).
\end{equation}
The agent can learn either a policy $\pi(a|s)$ without conditioning the latent skill variable when maximizing a knowledge-based or data-based intrinsic objective, or a latent-conditioned policy $\pi(a|s,z)$ when maximizing a competence-based intrinsic objective.
Common evaluation metrics for RFPT tasks include state coverage, skill diversity, and fine-tuning performance in downstream tasks.

On the contrary, in EIM tasks, the goal of the agent is to complete a specific downstream task and maximize only the expected task rewards. The optimization objective of EIM is
\begin{equation}
	\label{eqn: Exploration}
	L_{k}^{\text{EIM}}(\pi) = J_\text{E}(d_\pi) + \tau_k J_\text{I}(d_\pi),
\end{equation}
where $\tau_k$ is the coefficient of the intrinsic objective. Since the agent does not need to discover diverse skills for a specific task, $J_\text{I}(d_\pi)$ in EIM tasks is commonly a knowledge-based or data-based intrinsic objective without conditioning the latent skill variable. The evaluation metric for EIM is only the expected task rewards.
\begin{table*}[t]
	\centering
	\begin{tabular}{lll}
		\toprule
		\textbf{Algorithm} & \textbf{Intrinsic Objective} & \textbf{Intrinsic Reward} \\
		\midrule
		\textcolor{red}{ICM}~\cite{pathak2017curiosity} & $\E_{\rs} [ \rho_\pi^{-1}(s) ]$ & $\hat{\rho}_\pi^{-1}(s)$ \\ 
		\textcolor{red}{RND}~\cite{burda2018exploration} & $\E_{\rs} [ \rho_\pi^{-1}(s) ]$ & $\hat{\rho}_\pi^{-1}(s)$ \\ 
		\textcolor{red}{Dis.}~\cite{pathak2019self} & $\E_{\rs} [ \rho_\pi^{-1}(s) ]$ & $\hat{\rho}_\pi^{-1}(s)$ \\ 
		\textcolor{red}{MADE}~\cite{zhang2021made} & $\E_{\rs} [ (\rho_\pi^{-1}(s)d_\pi^{-1}(s))^{\nicefrac{1}{2}} ]$ & $(\hat{\rho}_\pi^{-1}(s)\hat{d}_\pi^{-1}(s))^{\nicefrac{1}{2}}$ \\ 
		\textcolor{red}{AGAC}~\cite{flet2021adversarially} & $\E_{\rs} [ D_{\text{KL}}(\pi(s)|\pi^\alpha(s)) ]$ & $\KL(\pi(s)|\pi^\alpha(s))$ \\ 
		\midrule
		\textcolor{green}{MaxEnt}~\cite{hazan2019provably} & $H(\rs)$ & $-\log\hat{d}_{\pi}(s)$ \\ 
		\textcolor{green}{APT}~\cite{liu2021behavior} & $H(\rs)$ & $-\log\hat{d}_{\pi}(f(s))$ \\ 
		\textcolor{green}{RE3}~\cite{seo2021state} & $H(\rs)$ & $-\log\hat{d}_{\pi}(f(s))$ \\ 
		\midrule
		\textcolor{blue}{VIC}~\cite{gregor2016variational} & $H(\rz)-H(\rz|\rs_f)$ & $\log q(z|s_f)$ \\ 
		\textcolor{blue}{DIAYN}~\cite{eysenbach2018diversity} & $H(\rz)-H(\rz|\rs)+H(\ra|\rs,\rz)$ & $\log q(z|s)$ \\ 
		\textcolor{blue}{VISR}~\cite{hansen2019fast} & $H(\rz)-H(\rz|\rs)$ & $S_c(\phi(s),z)$ \\ 
		\textcolor{blue}{DADS}~\cite{sharma2019dynamics} & $H(\rs'|\rs)-H(\rs'|\rs,\rz)$ & $-\log \hat{q}(s'|s)+\log q(s'|s,z)$ \\ 
		\textcolor{blue}{APS}~\cite{liu2021aps} & $H(\phi(\rs))-H(\phi(\rs)|\rz)$ & $-\log\hat{d}_{\pi}(\phi(s))+S_c(\phi(s),z)$ \\ 
		\textcolor{blue}{CIC}~\cite{laskin2022cic} & $H(\phi(\rs)),\ \text{s.t.}\ \phi\in\arg\min L^{\text{CIC}}(\phi(s),z)$ & $-\log\hat{d}_{\pi}(\phi(s))$ \\ 
		\textcolor{blue}{MOSS}~\cite{zhao2022mixture} & $\E_{\rm\sim \gB}(1-2m)H(\phi(\rs)|m)$ & $-(1-2m)\log\hat{d}_{\pi}(\phi(s))$ \\ 
		\textcolor{blue}{BeCL}~\cite{yang2023behavior} & $I(\rs;\rs^+),\ \text{s.t.}\ \phi\in\arg\min L^{\text{BeCL}}(\phi(s),z)$ & $\exp(-l^\text{BeCL})$ \\ 
		\textcolor{blue}{LSD}~\cite{park2021lipschitz} & $\E_{\rz,\rs} (\phi(s')-\phi(s))^Tz,\ \text{s.t.}\ \phi\in\arg\min L^{\text{LSD}}(\phi(s),z)$ & $(\phi(s')-\phi(s))^Tz$ \\ 
		\textcolor{blue}{CSD}~\cite{park2023controllability} & $\E_{\rz,\rs} (\phi(s')-\phi(s))^Tz,\ \text{s.t.}\ \phi\in\arg\min L^{\text{CSD}}(\phi(s),z)$ & $(\phi(s')-\phi(s))^Tz$ \\ 
		\midrule
		\textbf{\textcolor{blue}{CIM} (ours)} & $H(\phi(\rs)|\rz),\quad\text{s.t.}\ \phi\in\arg\min L_a(\phi(s),z)$ & $-\log\hat{d}_{\pi}(\phi(s)^Tz|z)$  \\
		\bottomrule
	\end{tabular}
	\caption{A summarization of IM algorithms. We denote knowledge-based, data-based, and competence-based IM methods in red, green, and blue, respectively. 1) In knowledge-based IM methods (in red), $\rho_\pi$ is the policy cover, $\hat{\rho}_pi$ is the estimated policy cover, $d_\pi$ is the state distribution, $\hat{d}$ is the estimated state distribution, $D_{\text{KL}}$ is the KL-divergence, and $\pi^\alpha$ is an adversarial policy in AGAC. 2) In data-based IM methods (in green), $H(\rs)$ stands for the entropy of the state distribution, and $f$ stands for an image encoder in pixel-based tasks and an identity encoder when in state-based tasks. 3) In competence-based IM methods (in blue), $z$ is the latent skill variable, $q$ is the discriminator, $\hat{q}$ is the estimated probability density, $s_f$ is the final state of the episode, $S_c$ is the cosine similarity between two vectors,  $\phi$ is the state encoder network, $s'\sim P(s'|s,a)$ is the subsequent state transitioned from the current state $s$ when action $a$ is taken, $m$ is a Bernoulli variable, $s^+$ is the positive sample in the contrastive objective.}
	\label{tab: IM algorithms}
\end{table*}

\subsection{Intrinsic Motivation Methods}
\label{subsec: IM}
We conduct a comprehensive comparison between our proposed CIM for RFPT and eighteen IM algorithms in regards of the intrinsic objective and the corresponding intrinsic reward function in \Cref{tab: IM algorithms}, including Intrinic Curiosity Module (ICM)~\cite{pathak2017curiosity}, Random Network Distillation (RND)~\cite{burda2018exploration}, Disagreement (Dis.)~\cite{pathak2019self}, MAximizing the DEviation from explored regions (MADE)~\cite{zhang2021made}, Adversarially Guided Actor-Critic (AGAC)~\cite{flet2021adversarially}, Maximum Entropy exploration (MaxEnt)~\cite{hazan2019provably}, Active Pre-Training (APT)~\cite{liu2021behavior}, Random Encoders for Efficient Exploration (RE3)~\cite{seo2021state}, Variational Intrinsic Control (VIC)~\cite{gregor2016variational}, Diversity Is All You Need (DIAYN)~\cite{eysenbach2018diversity}, Variational Intrinsic Successor featuRes (VISR)~\cite{hansen2019fast}, Dynamics-Aware Discovery of Skills (DADS)~\cite{sharma2019dynamics}, Active Pretraining with Successor features (APS)~\cite{liu2021aps}, Contrastive Intrinsic Control (CIC)~\cite{laskin2022cic}, Mixture Of SurpriseS (MOSS)~\cite{zhao2022mixture}, Behavior Contrastive Learning (BeCL)~\cite{yang2023behavior}, LSD, Controllability-Aware Skill Discovery (CSD)~\cite{park2023controllability}. Among these methods, ICM, RND, Dis., MADE, and AGAC belong to knowledge-based IM methods since their intrinsic objectives depend on the agent's all historical experiences. MaxEnt, APT, and RE3 fall under data-based IM methods, directly maximizing the state entropy. Meanwhile, VIC, DIAYN, VISR, DADS, APS, CIC, MOSS, BeCL, LSD, and CSD are classified as competence-based methods, all of which condition the latent skill variable.

\section{Constrained Intrinsic Motivation}
\label{sec: CIM}

In this section, we first present CIM for RFPT, a novel competence-based IM method that can learn dynamic and diverse skills efficiently. Specifically, we propose a constrained intrinsic objective $J_\text{I}^\text{CIM}$ for RFPT tasks, maximizing the conditional state entropy instead of the state entropy under a novel alignment constraint for the state representation. We then derive the corresponding intrinsic reward $r_\text{I}^\text{CIM}$ based on the Frank-Wolfe algorithm.
Secondly, we propose CIM for EIM to adaptively adjust the coefficient of the intrinsic objective in \Cref{eqn: Exploration} based on constrained policy optimization to mitigate the bias introduced by the intrinsic objective. We then derive the adaptive coefficient $\tau_k^\text{CIM}$ based on the Lagrangian duality theory.

\subsection{Constrained Intrinsic Motivation for RFPT}

In this section, we develop CIM for RFPT, a novel constrained intrinsic objective for unsupervised RL. To better clarify the motivation for the design of the constrained intrinsic objective, we first review current coverage- and mutual information-based methods and analyze their limitations.

\subsubsection{Problems of Previous Intrinsic Motivation Methods}

\begin{figure*}[t]
	\centering
	\begin{subcaptionblock}{\linewidth}
		\centering
		\includegraphics[width=0.115\linewidth]{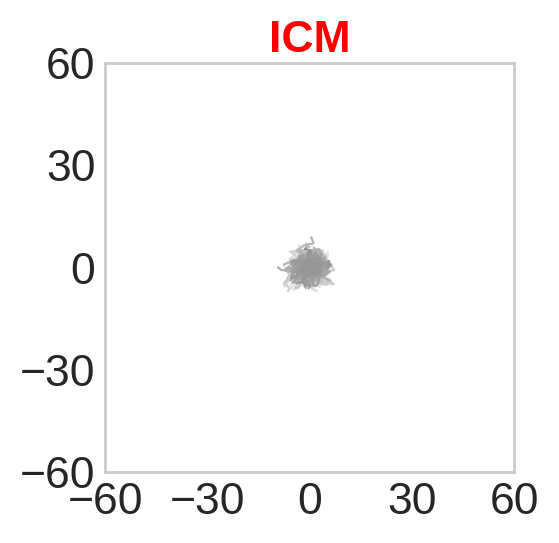}
		\includegraphics[width=0.115\linewidth]{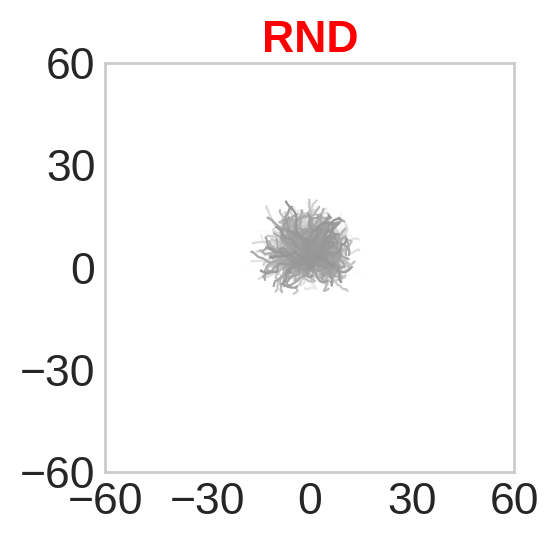}
		\includegraphics[width=0.115\linewidth]{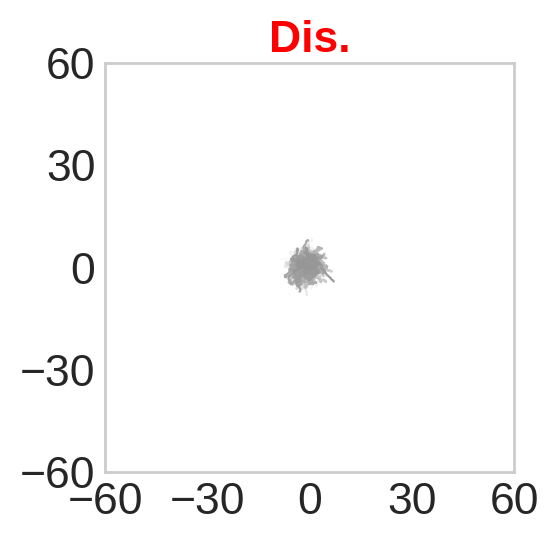}
		\includegraphics[width=0.115\linewidth]{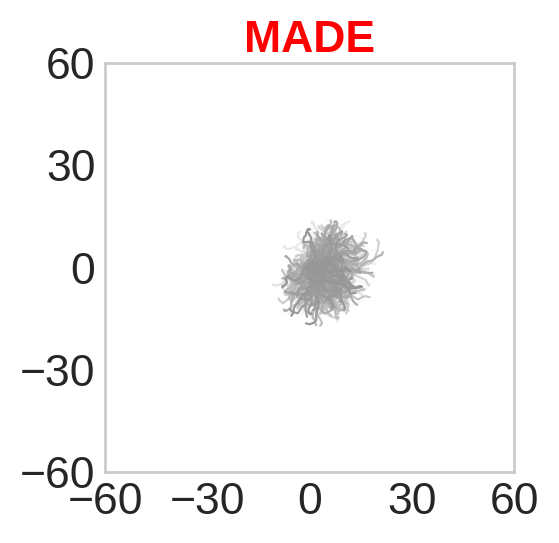}
		\includegraphics[width=0.115\linewidth]{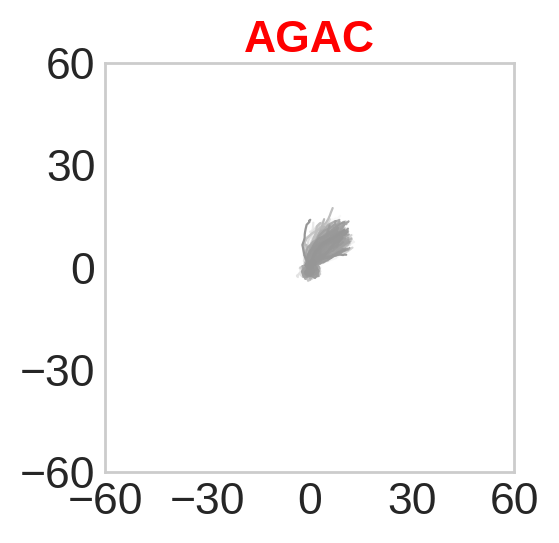}
		\includegraphics[width=0.115\linewidth]{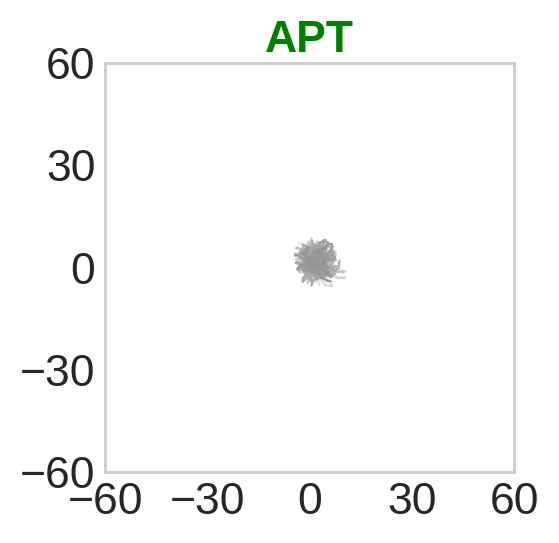}
		\includegraphics[width=0.115\linewidth]{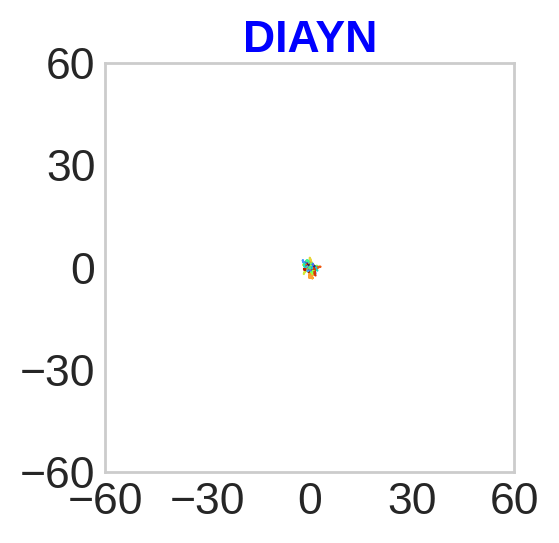}
		\includegraphics[width=0.115\linewidth]{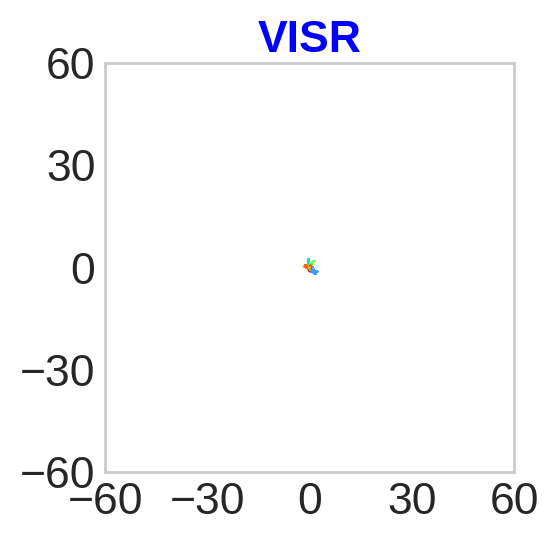}
		\includegraphics[width=0.115\linewidth]{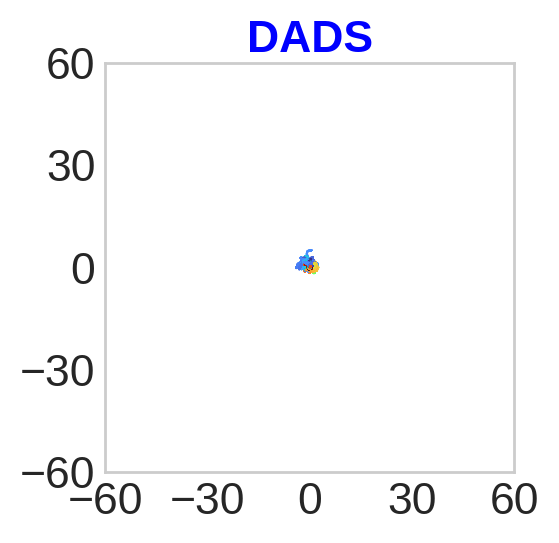}
		\includegraphics[width=0.115\linewidth]{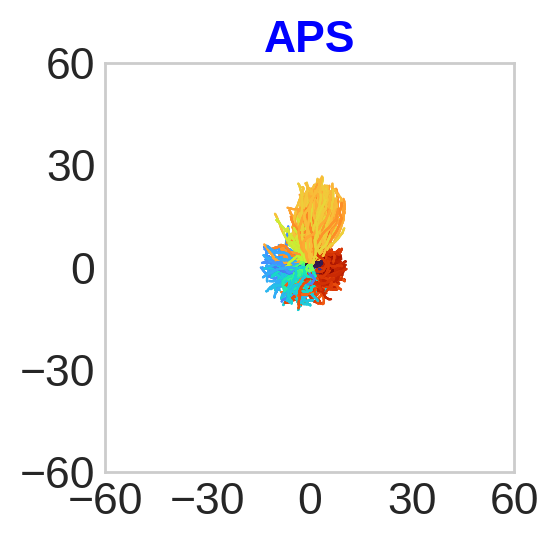}
		\includegraphics[width=0.115\linewidth]{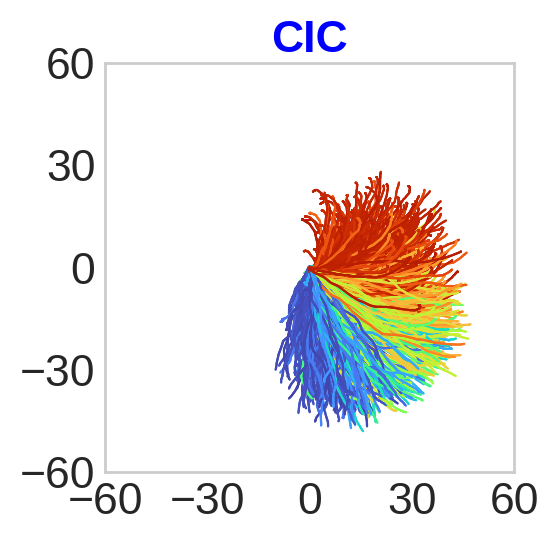}
		\includegraphics[width=0.115\linewidth]{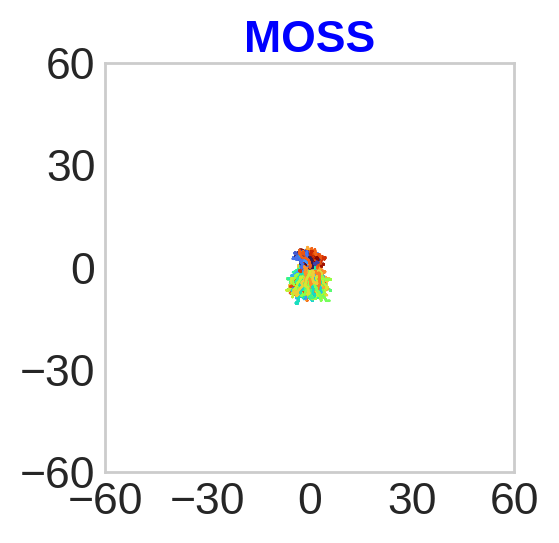}
		\includegraphics[width=0.115\linewidth]{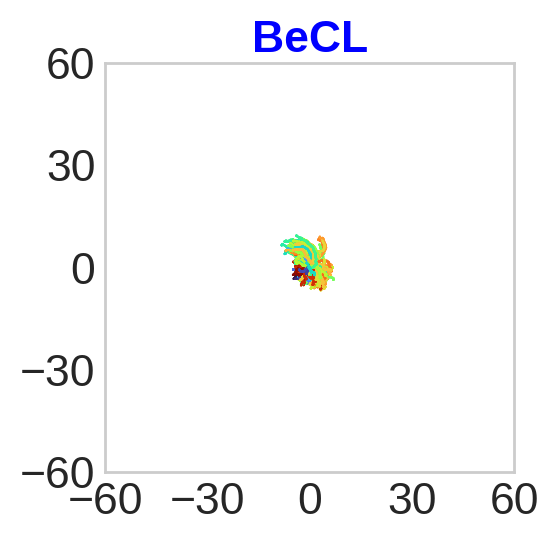}
		\includegraphics[width=0.115\linewidth]{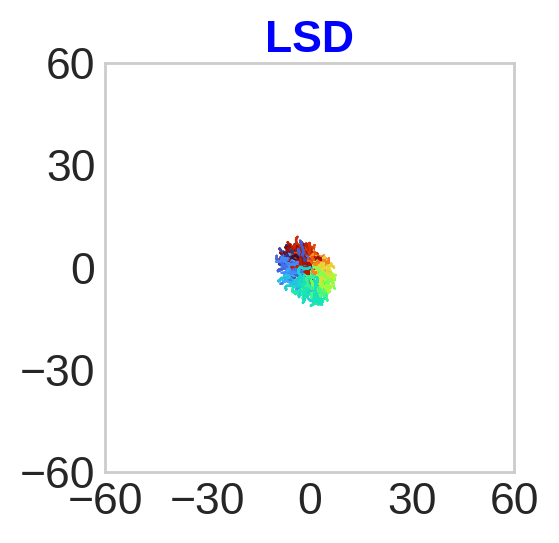}
		\includegraphics[width=0.115\linewidth]{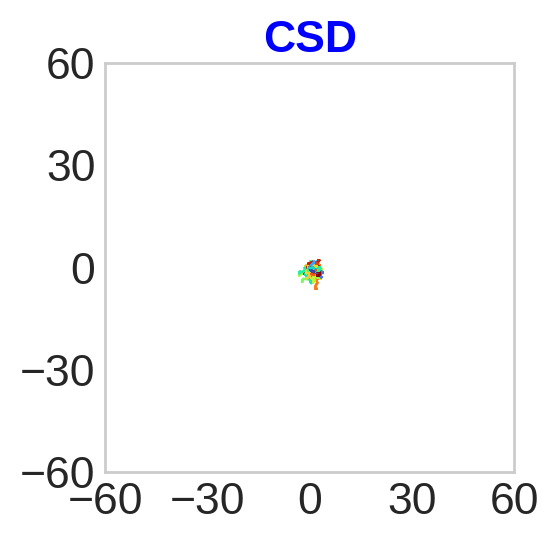}
		\includegraphics[width=0.115\linewidth]{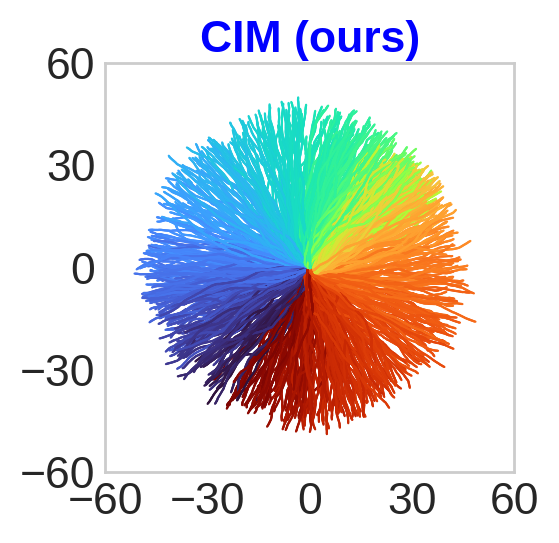}
	\end{subcaptionblock}
	\caption{Visualization of 2D continuous locomotion skills in Ant. Each color of the trajectories in competence-based IM methods (in blue) represents the direction of the latent skill variable $z$.}
	\label{fig: visualization in Ant}
\end{figure*}

Though knowledge-based and data-based IM methods may perform well in terms of state coverage in certain RFPT tasks, these methods lack awareness of latent skill variables and suffer from poor fine-tuning efficiency. To improve the fine-tuning performance in RFPT tasks, learning a latent-conditioned policy is necessary. However, existing competence-based IM methods perform poorly regarding skill diversity, state coverage, and sample inefficiency. We conjecture that there are two main issues:

\paragraph{Intrinsic objective.} Maximizing only the mutual information is not suitable for dynamic and diverse skill discovery in RFPT tasks. Recall the two types of decomposition for the mutual information, that is, $I(\rs;\rz)=H(\rz)-H(\rz|\rs)=H(\rs)-H(\rs|\rz)$. Note that minimizing $H(\rz|\rs)$ can be achieved with slight differences in states, and minimizing $H(\rs|\rz)$ clearly impedes the maximization of $H(\rs)$. Thus, neither can encourage the agent to cover large state space. Moreover, using $H(\rs)$ directly as the intrinsic objective (e.g., CIC and MOSS) also leads to low state coverage, as shown in \Cref{fig: visualization in Ant}. One of the key drawbacks of maximizing $H(\rs)$ is that it is challenging to estimate state density in the high-dimensional space.

\paragraph{Alignment constraint.} Current alignment constraints for the state encoder network are not efficient enough. For instance, LSD can learn dynamic skills with Lipschitz constraint on the state encoder network but suffers from heavy sample inefficiency. On the other hand, CIC applies noise contrastive estimation to formulate the alignment constraint on the state encoder network but fails to learn sufficiently dynamic and diverse skills.

\subsubsection{Design of Constrained Intrinsic Objective}
\label{subsubsec: alignment loss design}

To address the first issue, we propose choosing the conditional state entropy $H(\phi(\rs)|\rz)$ as the intrinsic objective. This is a key difference between our method and previous IM methods. Intuitively, by maximizing $H(\phi(\rs)|\rz)$, distances between adjacent states within the trajectories sampled by one skill are enlarged, which indicates a more \emph{dynamic} skill.

For the second issue, we propose maximizing a novel lower bound of the mutual information between the state representation and the latent skill variable to make the trajectories sampled by different skills \emph{distinguishable}, that is,
\begin{equation}
\label{eqn: alignment_loss_cim}
	\begin{aligned}
		I(\phi(\rs);\rz) &\ge \log N - L_a(\phi(s),z), \\
	\end{aligned}
\end{equation}
where $L_a(\phi(s),z)$ is the alignment loss as follows:
\begin{equation}
\label{eqn: alignment_loss_cim_2}
	\begin{aligned}
		L_a(\phi(s),z) &= \sum_i l_i^\text{CIM}, \\
		l_i^\text{CIM} &= -\phi^\text{diff}(\tau_i)^Tz_i + \\
		 &\log\sum\nolimits_{\tau_j\in S^-\bigcup\{\tau_i\}}\exp\left(\phi^\text{diff}(\tau_j))^Tz_i\right), \\
		 \phi^\text{diff}(\tau) &= \phi(s') - \phi(s),
	\end{aligned}
\end{equation}
$N$ is the total number of samples for estimating the mutual information, $\tau=(s,s')$ is the slice of a trajectory, and $S^-$ is a set of negative samples that contains trajectories sampled via skills other than $z_i$. We derive this lower bound based on Contrastive Predictive Coding~\cite{oord2018representation} by regarding the latent skill $z$ as the context and $\phi^\text{diff}(\tau)$ as the predictive coding. Based on \Cref{eqn: alignment_loss_cim}, the alignment constraint on the state encoder network $\phi(s)$ is 
\begin{equation}
	L_a(\phi(s),z) \le C,
\end{equation}
where $C$ is a constant. Theoretically, as indicated in Table 1, $C$ should represent the minimum of $L_a(\phi(s),z)$. In practice, we do not need to know the exact value of $C$. Instead, at each policy iteration step, we take several stochastic gradient descent steps on the alignment loss $L_a(\phi(s),z)$ to maximize the mutual information between the state representation $\phi(s)$ and the latent skill $z$.

The complete constrained intrinsic objective of CIM for RFPT is thus
\begin{equation}
\label{eqn: CIM for RFPT}
\begin{aligned}
	&\max_\pi J_\text{I}^{\text{CIM}}(d_{\pi}(\phi(s))) = H(\phi(\rs)|\rz) \\
	&\ \text{s.t.}\quad L_a(\phi(s),z) \le C,
\end{aligned}
\end{equation}
where $H(\phi(\rs)|\rz)$ is the conditional state entropy estimated in the state projection space, which depends on both the latent-conditioned policy network $\pi(\cdot|s,z)$ and the state encoder network $\phi(s)$, $d_{\pi}(\phi(s))$ is the distribution of the latent state $\phi(s)$ induced by the latent-conditioned policy $\pi(\cdot|s,z)$.

Interpreting $L_a(\phi(s),z)$ as an alignment loss provides us a novel insight to unify former competence-based IM methods. We can derive the alignment loss $l_i$ of all previous competence-based IM methods listed in \Cref{tab: IM algorithms}, e.g.,
\begin{equation}
\label{eqn: alignment loss}
	\begin{aligned}
		l_i^\text{MSE} &= \|\phi(s_i')-z_i\|_2^2, \\
		l_i^\text{vMF} &= -S_\text{c}(\phi(s_i'),z_i),\\
		l_i^\text{LSD} &= -\phi^\text{diff}(\tau_i)^Tz_i + \lambda(\|\phi^\text{diff}(\tau_i)\|-d(s,s')), \\
		l_i^\text{CIC} &= -S_\text{c} \bigl( \phi(\tau_i),\phi_z(z_i) \bigr) + \\
		&\log\sum\nolimits_{\tau_j\in S^-\bigcup\{\tau_i\}} \exp\Bigl( S_\text{c} \bigl( \phi(\tau_j),\phi_z(z_i) \bigr) \Bigr) , \\
		l_i^\text{BeCL} &= -S_\text{c} \bigl( \phi(s_i^+),\phi(s_i) \bigr) + \\
		&\log\sum\nolimits_{s_j\in S^-\bigcup\{s_i^{+}\}} \exp\Bigl( S_\text{c} \bigl( \phi(s_j),\phi(s_i) \bigr) \Bigr), \\
	\end{aligned}
\end{equation}
where $d(s,s')$ in $l_i^\text{LSD}$ is the state distance function, $S_\text{c}$ in $l_i^\text{vMF}$ is the cosine similarity between two vectors, $\phi_z$ in $l_i^\text{CIC}$ is a projection network for the latent skill vector, and $\phi(s_i^+)$ in $l_i^\text{BeCL}$ is a state representation from a certain skill as the positive sample while all others are negative samples.

\begin{figure*}[t]
	\centering
	\begin{subcaptionblock}{\linewidth}
		\centering
		\includegraphics[width=0.115\linewidth]{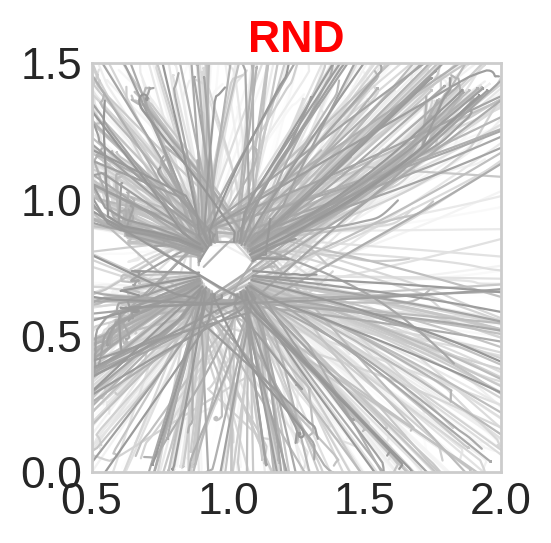}
		\includegraphics[width=0.115\linewidth]{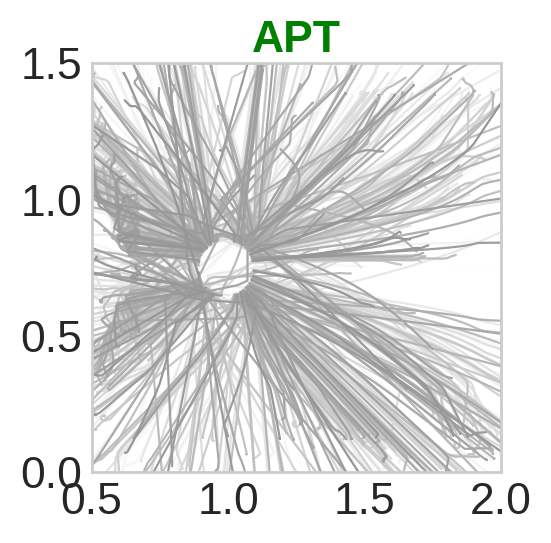}
		\includegraphics[width=0.115\linewidth]{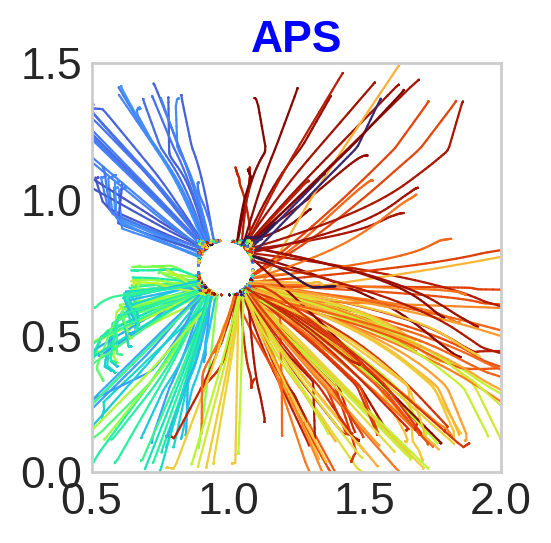}
		\includegraphics[width=0.115\linewidth]{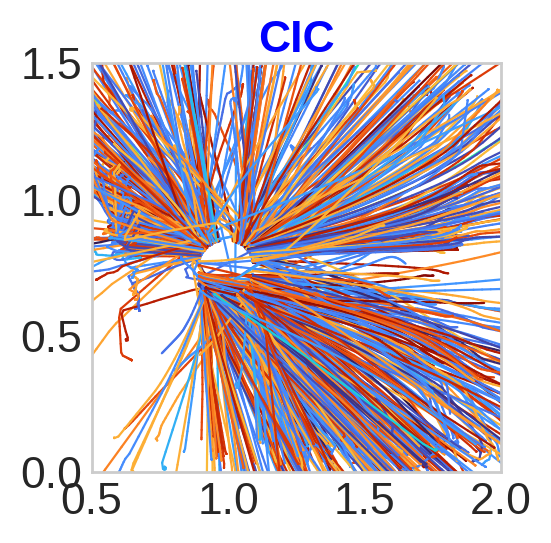}
		\includegraphics[width=0.115\linewidth]{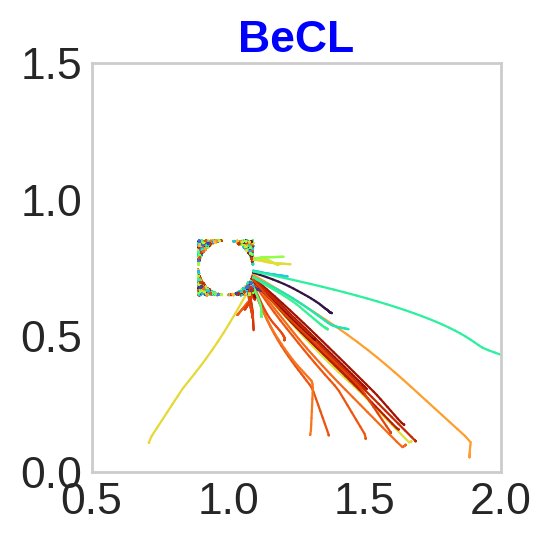}
		\includegraphics[width=0.115\linewidth]{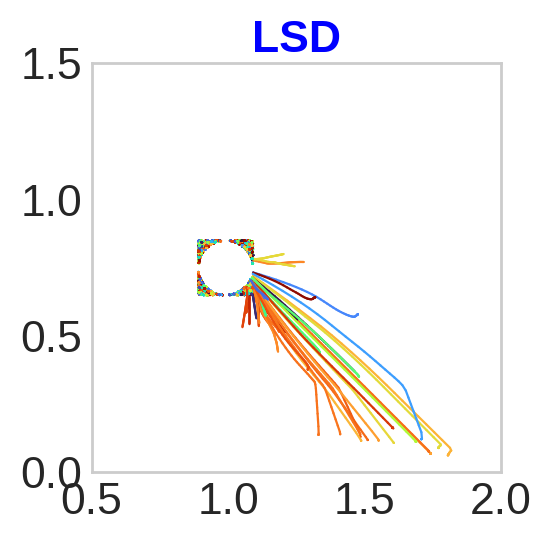}
		\includegraphics[width=0.115\linewidth]{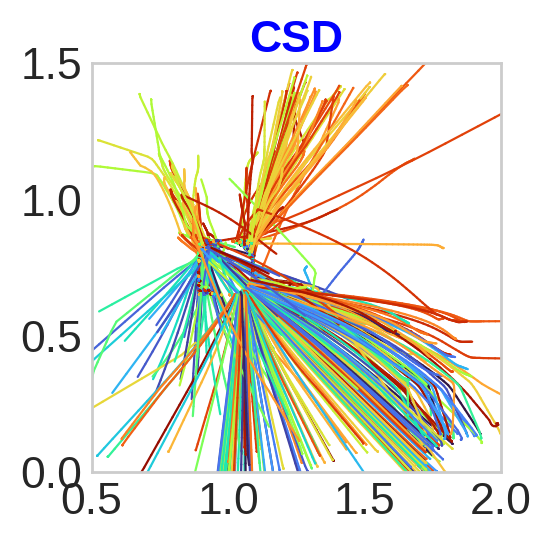}
		\includegraphics[width=0.115\linewidth]{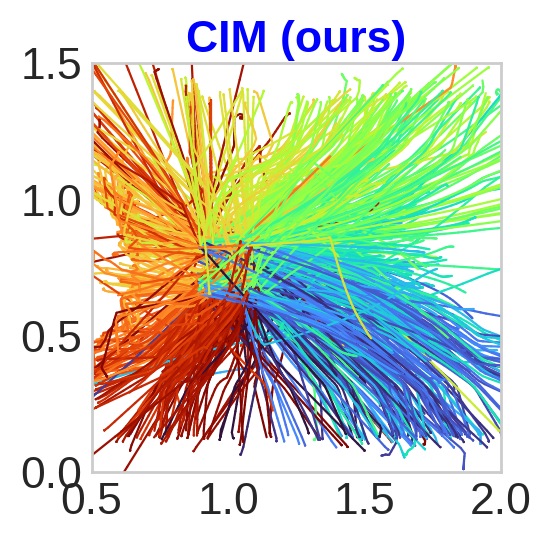}
	\end{subcaptionblock}
	\caption{Visualization of 2D continuous manipulation skills discovered by various IM methods in FetchSlide. Each color of the trajectories in competence-based IM methods (in blue) represents the direction of the latent skill variable $z$.}
	\label{fig: visualization in FetchSlide}
\end{figure*}

\subsubsection{Estimation of Conditional State Entropy}

We now introduce how to estimate the conditional state entropy $H(\phi(\rs)|\rz)$ involved in \Cref{eqn: CIM for RFPT}. Recall the definition of the conditional state entropy
\begin{equation}
\begin{aligned}
	H(\phi(\rs)|\rz) &= \E_{z\sim p_z} \left[ H(\phi(\rs)|\rz=z) \right] \\
	&= \E_{z\sim p_z} \E_{\phi(\rs)\sim d_{\pi}} \left[ -\log d_{\pi}(\phi(s)) \right].
\end{aligned}
\end{equation}
To estimate the outer expectation, we randomly sample the latent skill variables $z$ from a prior distribution $p_z(z)$. For discrete skills, $p_z(z)$ can be a categorical distribution $\text{Cat}(K,\vp)$ that is parameterized by $\vp$ over a size-$K$ the sample space, where $\evp_i$ denotes the probability of the $i-$th skill. For continuous skills,  we can select $p(z)$ as a uniform distribution $\gU^{n_z}(a,b)$ over the interval $[a,b]$, where $n_z$ is the dimension of the skill.

To estimate the inner expectation, we roll out trajectories using the latent-conditioned policy $\pi(\cdot|s,z)$ with $z$ fixed. During the sampling phase, $z$ is randomly sampled at the beginning of each episode and remains fixed throughout the entire trajectory. We store the state-skill pair ($s$,$z$) in the replay buffer. During the training phase, for each pair ($s$,$z$), we concatenate them as $[s,z]$ to be the input of the latent-conditioned policy $\pi(\cdot|s,z)$.

To estimate the state density $d_{\pi}$, instead of training a parameterized generative model, we leverage a more practical non-parametric $\xi-$nearest neighbor ($\xi-$NN) estimator
\begin{equation}
	\hat{d}_{\pi}(s_i) = \frac{1}{\lambda\left(B_\xi(s_{i})\right)}\int_{B_\xi(s_{i})} d_{\pi}(s) \mathrm{d}s,
\end{equation}
where $\lambda$ is the Lebesgue measure on $\mathbb{R}^d$, $B_\xi$ is the smallest ball centered on $s_i$ containing its $\xi$-th nearest neighbour $s_i^\xi$.

\paragraph{Lower bound of conditional state entropy.} Each skill can be stochastic if we directly maximize the conditional state entropy $H(\rs|\rz)$. To address this, we propose maximizing \textit{the lower bound of $H(\rs|\rz)$} to encourage the skill $z$ to produce large state variations along the direction of $z$ in the latent space instead of being fully stochastic. To derive the lower bound of $H(\rs|\rz)$, we first define a projection function $g_z(\phi(s))=\phi(s)^Tz$ for a fixed skill $z$. It is easy to verify that $H(\phi(\rs)|\rz)\ge H(g_z(\phi(\rs))|\rz))$ with equality iff $S_c(\phi(s),z)=1$, that is, $H(g_z(\phi(\rs))|\rz))$ is a lower bound of $H(\phi(\rs)|\rz)$. We thus can maximize $H(g_z(\phi(\rs))|\rz))$ to maximize $H(\phi(\rs)|\rz)$ and estimate the distribution of the \textit{one-dimensional} random variable $g_z(\phi(s))$ for each $z$. 

\paragraph{Intrinsic reward.} Based on the above design, we can derive the intrinsic reward of CIM for RFPT as $r_\text{I}^\text{CIM}(s) = \log\|g_z(\phi(s))-g_z(\phi(s))^\xi\|$. Here, $g_z(\phi(s))^\xi$ means the $\xi$-th nearest neighbor of $g_z(\phi(s))$. We adopted an average-distance version similar to APT to make training more stable:
\begin{equation}
\label{eqn: rew_cim}
	r_\text{I}^\text{CIM}(s) = \log\left(1 + \frac{1}{\xi}\sum_{j=1}^{\xi}\|g_z(\phi(s))-g_z(\phi(s))^j\|\right).
\end{equation}
Intuitively, $r_\text{I}^\text{CIM}(s)$ measures how sparse the state $s$ is in the projection subspace spanned by its corresponding latent skill $z$. This reward function can be justified based on the procedure of the Frank-Wolfe algorithm. Specifically, since $L^\text{RFPT}_k$ is concave in $d_\pi$, maximizing $L^\text{RFPT}_k$ involves solving $d_{\pi_{k+1}}\in\arg\max\langle \nabla_{d_\pi} L(d_{\pi_k}), d_{\pi_k} - d_\pi\rangle$ iteratively~\cite{hazan2019provably}. This iterative step is equivalent to policy optimization using a reward function proportional to $\nabla_{d_\pi} L(d_{\pi_k})$.

\subsection{Constrained Intrinsic Motivation for EIM}

In this section, we present our CIM for EIM, an adaptive coefficient for the intrinsic objective in \Cref{eqn: Exploration}. Currently, IM methods for EIM tasks commonly use a constant coefficient or an exponentially decaying coefficient, which requires costly hyperparameter tunning. To avoid this, we propose reformulating \Cref{eqn: Exploration} by regarding the extrinsic objective as a constraint for the intrinsic objective, i.e.,
\begin{equation}
\label{eqn: CIM for EIM}
	\max_{d_{\pi} \in \gK} J_\text{I}(d_\pi),\text{ s.t. } J_\text{E}(d_\pi) \ge R_k,
\end{equation}
where $R_k$ represents the expected reward at the $k$-th iteration of policy optimization. We approximate $R_k$ via $\hat{R}_k = \max_{j\in\{1,2,...,k-1\}} J_\text{E}(d_{\pi_{j}})$. We then leverage the Lagrangian method to solve \Cref{eqn: CIM for EIM}. The corresponding Lagrangian dual problem is 
$\min_{\lambda\ge0}\max_{d_\pi} J_\text{I}(d_\pi) + \lambda_k (J_\text{E}(d_{\pi}) -  \hat{R}_k)$.
The Lagrangian multiplier $\lambda$ is updated by Stochastic Gradient Descent (SGD), that is, $\lambda_k = \lambda_{k-1} - \eta (J_\text{E}(d_{\pi_{k}}) -  \hat{R}_{k-1})$ where $\eta$ is the updating step size of $\lambda_k$. Observing that $\gL_k(d_\pi,\lambda_k) \propto J_\text{E}(d_\pi) + \lambda_k^{-1}J_\text{I}(d_\pi)$, the adaptive coefficient $\tau_k^\text{CIM}$ is then derived as
\begin{equation}
	\tau_k^\text{CIM}=\min\{\{\lambda_{k-1}-\eta (J_\text{E}(d_{\pi_{k}})-\hat{R}_{k-1})\}^{-1},1\},
\end{equation}
where the outer minimization is to ensure numerical stability. As the Lagrangian multiplier $\lambda_k$ grows, the penalty for the violation of  $\tau_k^\text{CIM}$ gradually tends to zero; that is, the bias introduced by the intrinsic objective $J_\text{I}$ is adaptively reduced.

\begin{table*}[t]
	\centering
	\resizebox{0.99\textwidth}{!}{
	\begin{tabular}{lCCCCCCCC}
		\toprule
		\textbf{Environment} & \textcolor{red}{\textbf{RND}} & \textcolor{green}{\textbf{APT}} & \textcolor{blue}{\textbf{APS}} & \textcolor{blue}{\textbf{CIC}} & \textcolor{blue}{\textbf{LSD}} & \textcolor{blue}{\textbf{CSD}} & \textbf{\textcolor{blue}{CIM} (ours)}\\
		\midrule
		Ant (29D) & 123\p{15} & 33\p3 & 192\p75 & 697\p200 & 50\p24 & 4\p0 & \textbf{1042}\p\textbf{158} \\
		Humanoid (378D) & 22\p1 & 22\p1 & 107\p33 & 64\p11 & 8\p1 & 4\p0 & \textbf{1135}\p\textbf{360} \\
		\midrule
		FetchPush (25D) & 137\p22 & \textbf{154}\p{17} & 79\p14 & 150\p34 & 24\p12 & 105\p48 & 141\p15 \\
		FetchSlide (25D) & 182\p52 & 185\p49 & 178\p33 & \textbf{223}\p\textbf{3} & 31\p33 & 114\p79 & 187\p16 \\
		\bottomrule
	\end{tabular}}
	\caption{State coverage of 2D continuous locomotion or manipulation skills discovered by various typical IM methods. We denote knowledge-based, data-based, and competence-based IM methods in red, green, and blue, respectively.}
	\label{tab: state coverage}
\end{table*}

\begin{table*}[t]
	\centering
	\resizebox{\textwidth}{!}{
	\begin{tabular}{lCCCCCCCC}
		\toprule
		\textbf{Task} & \textbf{DDPG} & \textcolor{red}{\textbf{RND}} & \textcolor{green}{\textbf{Proto}} & \textcolor{blue}{\textbf{APS}} & \textcolor{blue}{\textbf{CIC}} & \textcolor{blue}{\textbf{MOSS}} & \textcolor{blue}{\textbf{BeCL}} & \textbf{\textcolor{blue}{CIM} (ours)}\\
		\midrule
		Flip & 536\p66 & 470\p47 & 523\p89 & 407\p104 & \textbf{709}\p\textbf{172} & 425\p77 & 628\p46 & 664\p80 \\
		Run & 274\p22 & 403\p105 & 347\p102 & 128\p38 & 492\p81 & 244\p13 & 467\p81 & \textbf{585}\p\textbf{27} \\
		Stand & 931\p18 & 907\p16 & 861\p79 & 698\p215 & 939\p28 & 862\p100 & \textbf{951}\p\textbf{3} & 941\p21 \\
		Walk & 777\p89 & 844\p99 & 828\p70 & 577\p133 & 905\p22 & 684\p40 & 781\p221 & \textbf{921}\p\textbf{30}\\
		\midrule
		Score & 0.69\p0.23 & 0.72\p0.20 & 0.70\p0.20 & 0.49\p0.25 & 0.85\p0.18 & 0.60\p0.22 & 0.78\p0.19 & \textbf{0.86}\p\textbf{0.11}\\
		\bottomrule
	\end{tabular}}
	\caption{Fine-tuning performance (average episode rewards\p standard deviations) of eight typical methods in Walker domain of URLB. We report the normalized average score in the last row. We denote knowledge-based, data-based, and competence-based IM methods in red, green, and blue, respectively.}
	\label{tab: results in URLB}
\end{table*}

\begin{table}[t]
	\centering
	\resizebox{0.92\columnwidth}{!}{
	\begin{tabular}{lrrrrrr}
		\toprule
		\textbf{Environment} & $l_i^\text{MSE}$ & $l_i^\text{vMF}$ & $l_i^\text{LSD}$ & $l_i^\text{CIC}$ & $l_i^\text{BeCL}$ & $l_i^\text{CIM}$ \textbf{(ours)}\\
		\midrule
		Ant (29D) & 64 & 371 & 28 & 746 & 726 & \textbf{1042} \\
		\bottomrule
	\end{tabular}}
	\caption{State coverage when replacing $l_i^\text{CIM}$ in $L_a(\phi(s),z)$ with other alignment losses $l_i$ as listed in \Cref{eqn: alignment loss} in Ant.}
	\label{tab: ablation on alignment loss}
\end{table}

\begin{table}[t]
	\centering
	\resizebox{0.9\columnwidth}{!}{
	\begin{tabular}{lrrrr}
		\toprule
		\textbf{Environment} & $n_z=2$ & $n_z=3$ & $n_z=10$ & $n_z=64$ \\
		\midrule
		Ant (29D) & \textbf{1042}\p{158} & 875\p240 & 901\p20 & 615\p54\\
		\bottomrule
	\end{tabular}}
	\caption{State coverage when varying the skill dimension $n_z$ in Ant.}
	\label{tab: ablation on skill dimensions}
\end{table}

\begin{table}[t]
	\centering
	\resizebox{0.6\columnwidth}{!}{
	\begin{tabular}{lrrrr}
		\toprule
		\textbf{Coefficient} & \textbf{SHC} & \textbf{SA} & \textbf{SHS} & \textbf{SGW} \\
		\midrule
		$\tau_k^{\text{C}}$ & 0.27 & 0.74 & 0.18 & 0.01 \\
		$\tau_k^{\text{CIM}}$ \textbf{(ours)} & \textbf{1} & \textbf{0.97} & \textbf{1} & \textbf{1} \\
		\bottomrule
	\end{tabular}}
	\caption{Test-time average episode rewards using different coefficient schemes across four sparse-reward EIM tasks.}
	\label{tab: more exp for CIM for EIM}
\end{table}

\begin{figure}[t]
	\centering
	\subcaptionbox{\label{fig: maze-a}}
		{\includegraphics[width=0.44\columnwidth]{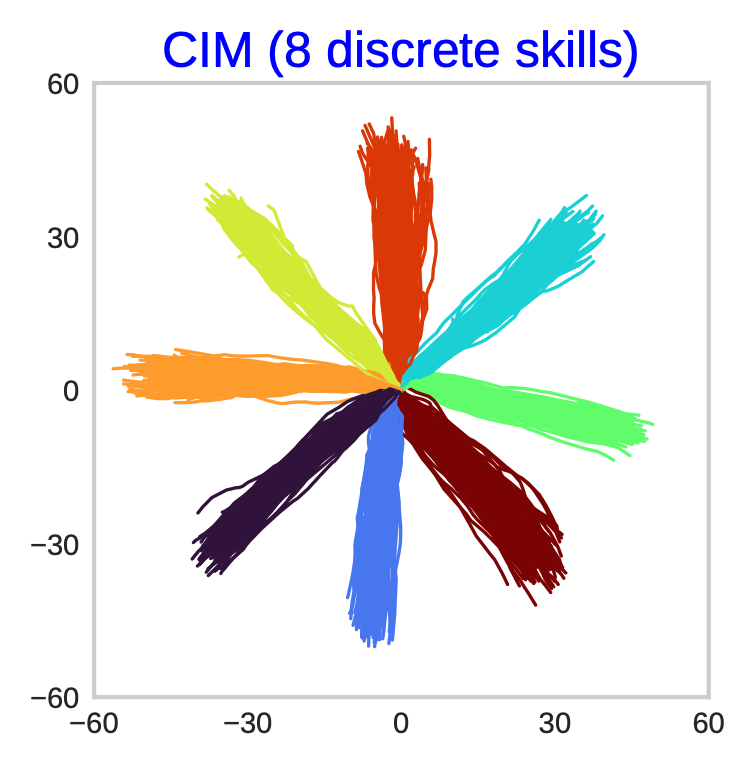}}
	\subcaptionbox{\label{fig: maze-b}}
		{\includegraphics[width=0.44\columnwidth]{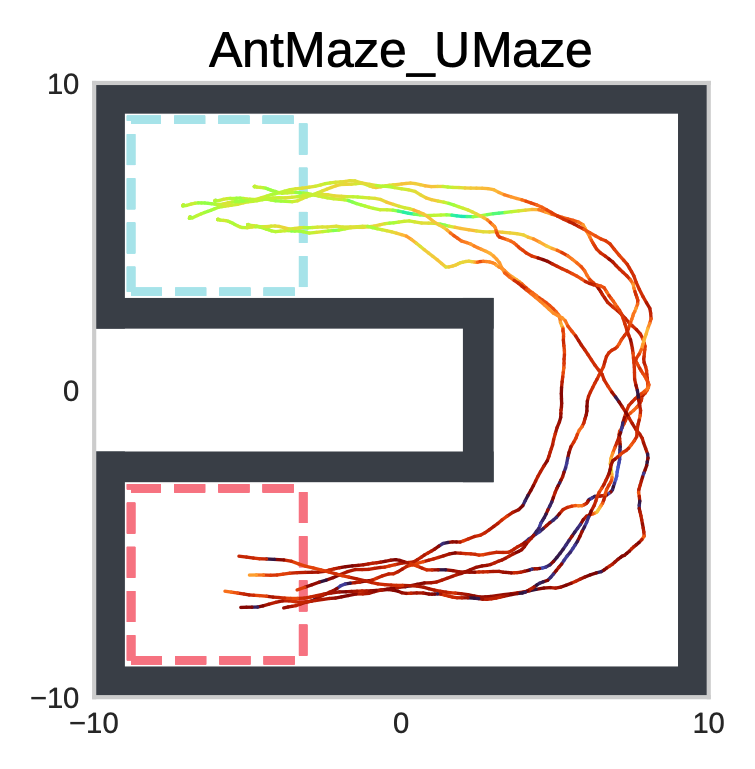}}
	\subcaptionbox{\label{fig: maze-c}}
		{\includegraphics[width=\columnwidth]{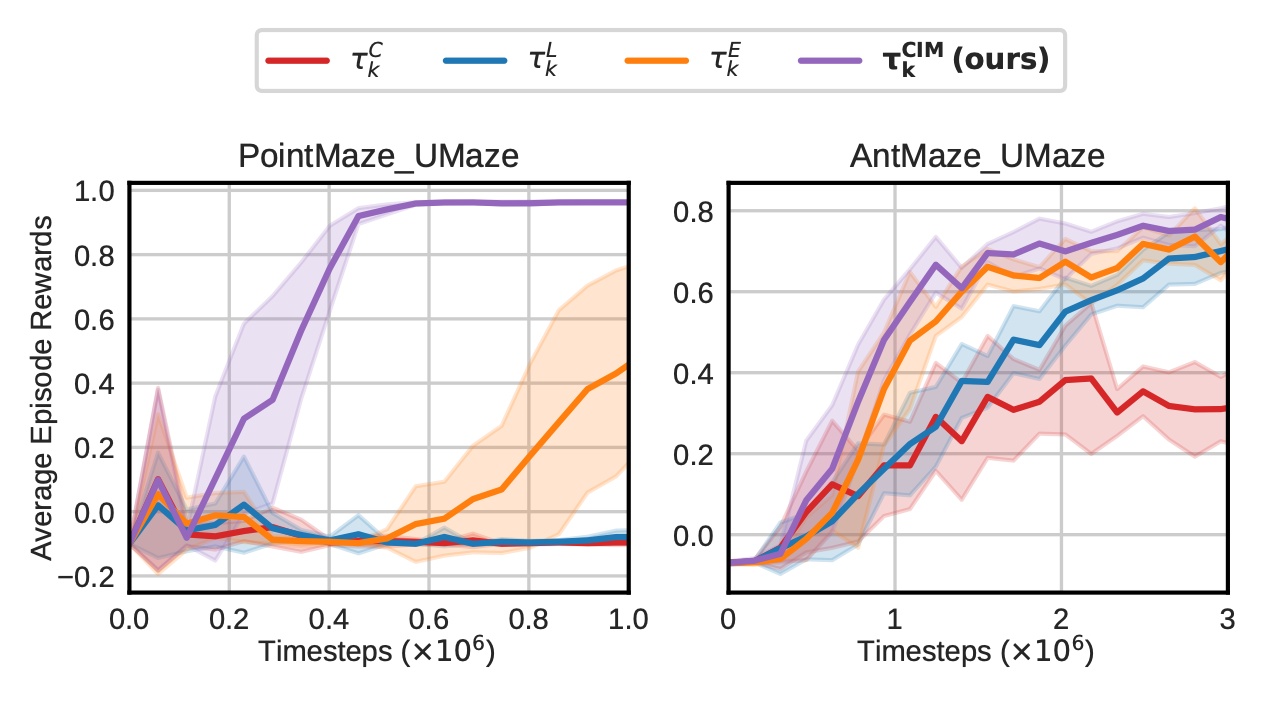}}
	\caption{\subref{fig: maze-a} Discrete CIM with $n_z=8$ in Ant. \subref{fig: maze-b} Trajectory visualization of the meta-controller where the color of each sub-trajectory reflects the direction of the skill. \subref{fig: maze-c} Learning curves using different coefficients of the intrinsic objective.
	}
	\label{fig: maze}
\end{figure}

\section{Experiments}
\label{sec: experiments}

\subsection{Experimental Setup}

\paragraph{Experimental setup for RFPT.} We evaluate our intrinsic bonus $r_\text{I}^\text{CIM}$ for RFPT tasks on four Gymnasium environments, including two locomotion environments (Ant and Humanoid) and two manipulation environments (FetchPush and FetchSlide). We compare CIM for RFPT with fifteen IM methods in \Cref{tab: IM algorithms}, including 1) four knowledge-based IM methods: ICM, RND, Dis., MADE, and AGAC; 2) one data-based IM method: APT; 3) and nine competence-based methods: DIAYN, VISR, DADS, APS, CIC, MOSS, BeCL, LSD, and CSD.

\paragraph{Experimental setup for EIM.} We evaluate our adaptive coefficient $\tau_k^\text{CIM}$ for EIM in two navigation tasks (PointMaze\_UMaze and AntMaze\_UMaze) in D4RL~\cite{fu2020d4rl}, and four sparse-reward tasks (SparseHalfCheetah, SparseAnt, SparseHumanoidStandup, and SparseGridWorld). $\tau_k^\text{CIM}$ is orthogonal with any intrinsic bonuses $r_\text{I}$. Unless otherwise mentioned, we adopt the state-of-the-art data-based intrinsic bonus $r_\text{I}^\text{APT}=\log(1+\nicefrac{1}{k}\sum_{j=1}^k\|\phi(s)-\phi(s)^j\|)$. The total instant reward is then $r=r_\text{E}+\tau_k^\text{CIM}r_\text{I}^\text{APT}$. We compare CIM for EIM with three baseline coefficient schemes, i.e., the constant coefficient $\tau_k^\text{C}\equiv 1$, the linearly decaying coefficient $\tau_k^\text{L}=(1-k/T)$, and the exponentially decaying coefficient $\tau_k^\text{E}=0.001^k$.

\subsection{Results in RFPT Tasks}

\paragraph{Visualization of skills.} As previous works like LSD do, we train CIM for RFPT to learn diverse locomotion continuous skills in the Ant and Humanoid environment and diverse manipulation skills in FetchPush and FetchSlide. The learned skills are visualized as trajectories of the agent on the $x-y$ plane in \Cref{fig: visualization in Ant} and \Cref{fig: visualization in FetchSlide}.
Our CIM for RFPT outperforms all 15 baselines in terms of skill diversity and state coverage. The skills learned via CIM are interpretable because of our alignment loss; the direction of the trajectory on the $x-y$ plane changes consistently with the change in the direction of the skill.
Specifically, CIM excels at learning dynamic skills that move far from the initial location in almost all possible directions, while most baseline methods fail to discover such diverse and dynamic primitives. Their trajectories are non-directional or less dynamic than CIM, especially in two locomotion tasks. Competence-based approaches like DIAYN, VISR, and DADS directly maximize the mutual information objectives but learn to take static postures instead of dynamic skills; such a phenomenon is also reported in LSD and CIC. Although APS and CIC can learn dynamic skills by directly maximizing the state entropy, CIM discovers skills that reach farther and are more interpretable via maximizing the lower bound of the state entropy. As for the two variants of CIC, MOSS and BeCL, they perform even worse than CIC in all tasks, reflecting their limitation in skill discovery. Lastly, LSD and CSD cannot learn dynamic skills within limited environment steps in Ant and Humanoid due to their low sample efficiency. Though they perform better in manipulation tasks than locomotion tasks, their learned skills are rambling compared with our CIM.

\paragraph{State coverage.} To make a quantitative comparison between various IM methods, we measure their state coverage. The state coverage in Ant and Humanoid is determined by calculating the number of 2.5 $\times$ 2.5 $\rm m^2$ bins occupied on the x-y plane, based on 1000 randomly sampled trajectories. This was then averaged over five runs. For FetchPush and FetchSlide, we use smaller bins. As shown in \Cref{tab: state coverage}, CIM significantly outperforms all the baseline methods in two torque-as-input locomotion tasks and is comparable in two position-as-input manipulation tasks. Although the state coverage of CIM is slightly lower than APT and CIC in FetchPush and FetchSlide, the skills learned via CIM are more interpretable, as shown in \Cref{fig: visualization in FetchSlide}.

\paragraph{Fine-tuning efficiency in URLB.} We also evaluate CIM for RFPT in URLB, a benchmark environment for RFPT in terms of fine-tuning efficiency. The results are presented in \Cref{tab: results in URLB}. The score (the last line of the table) is standardized by the performance of the expert DDPG, the same as in URLB and CIC. CIM performs better in Run and Walk tasks and achieves the highest average score. The dynamic skills learned through CIM for RFPT can be adapted quickly to diverse fine-tuning tasks, including flipping and standing. Our experiments also show that the skill dimension $n_z=3$ is better for CIM to discover flipping skills than $n_z=2$. The fixed skill selection mechanism for CIM is the same as CIC.

\paragraph{Ablation study.} According to the results in \Cref{tab: ablation on alignment loss}, loss functions that follow the NCE style, such as $l_i^\text{CIC}$, $l_i^\text{BeCL}$, and $l_i^\text{CIM}$, perform better than other styles like MSE and vMF. Besides, $l_i^\text{CIM}$ is the most effective.
As shown in \Cref{fig: maze-a}, our CIM can also be utilized to discover discrete diverse and dynamic skills, though it is mainly designed for continuous skills.
Moreover, our CIM for RFPT is also robust to the number of skill dimensions, as shown in \Cref{tab: ablation on skill dimensions}. Based on the ablation study, we can conclude that the two components of CIM, i.e., minimizing NCE-style alignment loss and maximizing conditional state entropy, are equally critical. Specifically, the results in \Cref{tab: ablation on alignment loss} show that replacing the alignment loss of CIM with a trivial MSE loss reduces the state coverage in Ant from 1042 to 64. Moreover, \Cref{tab: state coverage} reveals that the state coverage achieved by CIM can reach 1135 in the challenging 378-dimensional Humanoid, while that achieved by CIC and BeCL, which use similar NCE-style alignment losses, is lower than 100.

\subsection{Results in EIM Tasks}

In PointMaze, we directly train a policy to control the Point without learning low skills since the environment dynamics are simple. In AntMaze, we train a meta-controller on top of the latent-conditioned policy pre-trained via our CIM for RFPT method. The meta-controller observes the target goal concatenated to the state observation $[s;s_g]$ and outputs the skill latent variable $z$ at each timestep. We visualize the trajectories of the Ant in the $x-y$ plane as shown in \Cref{fig: maze-b}, where the skills in a single trajectory gradually change to make the Ant turn a corner. \Cref{fig: maze-c} shows that the Lagrangian-based adaptive coefficient $\tau_k^\text{CIM}$ outperforms three baseline coefficients, especially in PointMaze. Specifically, we can observe a small peak in the early stage of the training in PointMaze, which means the agent can reach the randomly generated target point with a small probability at the beginning. However, as the training processes, the agent is distracted by the intrinsic bonuses when using a trivial coefficient $\tau_k^\text{C}$ or $\tau_k^\text{L}$.
Moreover, other latent-conditioned policies are of poor quality, and we fail to train a mete-controller on top of those policies. We also conduct experiments to demonstrate the performance of CIM for EIM across four sparse-reward locomotion tasks. The results in \Cref{tab: more exp for CIM for EIM} indicate that $\tau_k^{\text{CIM}}$ can effectively reduce the bias introduced by intrinsic rewards, thereby enhancing test-time average episode rewards in EIM tasks.

\section{Conclusion}

In this paper, we proposed Constrained Intrinsic Motivation (CIM) for RFPT and EIM tasks, respectively. For RFPT tasks, we designed a novel constrained intrinsic objective to discover dynamic and diverse skills. For EIM tasks, we designed an adaptive coefficient $\tau_k^\text{CIM}$ for the intrinsic objective based on constrained policy optimization. Our experiments demonstrated that CIM for RFPT outperformed all fifteen baselines across various MuJoCo environments regarding diversity, state coverage, sample efficiency, and fine-tuning performance. The latent-conditioned policy learned via CIM for RFPT was successfully applied to solve complex EIM tasks via training a meta-controller on top of it. We also empirically verified the effectiveness of our adaptive coefficient $\tau_k^\text{CIM}$ in multiple EIM tasks.

\section*{Acknowledgments}

We thank the anonymous reviewers for their valuable feedback. This work was partially supported by HK RGC under Grants (CityU 11218322, R6021-20F, R1012-21, RFS2122-1S04, C2004-21G, C1029-22G, and N\_CityU139/21) and the National Natural Science Foundation of China (U21B2018, 62161160337, 61822309, U20B2049, 61773310, U1736205, 61802166, 62276067).

\bibliographystyle{named}
\bibliography{camera_ready.bib}

\begin{thebibliography}{}

\bibitem[\protect\citeauthoryear{Bai \bgroup \em et al.\egroup
  }{2021}]{bai2021principled}
Chenjia Bai, Lingxiao Wang, Lei Han, Jianye Hao, Animesh Garg, Peng Liu, and
  Zhaoran Wang.
\newblock Principled exploration via optimistic bootstrapping and backward
  induction.
\newblock In {\em Proc. of the International Conference on Machine Learning
  (ICML)}, pages 577--587, 2021.

\bibitem[\protect\citeauthoryear{Barto}{2013}]{barto2013intrinsic}
Andrew~G. Barto.
\newblock Intrinsic motivation and reinforcement learning.
\newblock In {\em Intrinsically Motivated Learning in Natural and Artificial
  Systems}, pages 17--47. 2013.

\bibitem[\protect\citeauthoryear{Bellemare \bgroup \em et al.\egroup
  }{2016}]{bellemare2016unifying}
Marc~G. Bellemare, Sriram Srinivasan, Georg Ostrovski, Tom Schaul, David
  Saxton, and R{\'{e}}mi Munos.
\newblock Unifying count-based exploration and intrinsic motivation.
\newblock In {\em Proc. of the Conference on Neural Information Processing
  Systems (NeurIPS)}, pages 1471--1479, 2016.

\bibitem[\protect\citeauthoryear{Burda \bgroup \em et al.\egroup
  }{2019}]{burda2018exploration}
Yuri Burda, Harrison Edwards, Amos~J. Storkey, and Oleg Klimov.
\newblock Exploration by random network distillation.
\newblock In {\em Proc. of the International Conference on Learning
  Representations (ICLR)}, 2019.

\bibitem[\protect\citeauthoryear{Chen \bgroup \em et al.\egroup
  }{2022}]{chen2022redeeming}
Eric Chen, Zhang{-}Wei Hong, Joni Pajarinen, and Pulkit Agrawal.
\newblock Redeeming intrinsic rewards via constrained optimization.
\newblock In {\em Proc. of the Conference on Neural Information Processing
  Systems (NeurIPS)}, pages 4996--5008, 2022.

\bibitem[\protect\citeauthoryear{Eysenbach \bgroup \em et al.\egroup
  }{2019}]{eysenbach2018diversity}
Benjamin Eysenbach, Abhishek Gupta, Julian Ibarz, and Sergey Levine.
\newblock Diversity is all you need: Learning skills without a reward function.
\newblock In {\em Proc. of the International Conference on Learning
  Representations (ICLR)}, 2019.

\bibitem[\protect\citeauthoryear{Flet{-}Berliac \bgroup \em et al.\egroup
  }{2021}]{flet2021adversarially}
Yannis Flet{-}Berliac, Johan Ferret, Olivier Pietquin, Philippe Preux, and
  Matthieu Geist.
\newblock Adversarially guided actor-critic.
\newblock In {\em Proc. of the International Conference on Learning
  Representations (ICLR)}, 2021.

\bibitem[\protect\citeauthoryear{Fu \bgroup \em et al.\egroup
  }{2017}]{fu2017ex2}
Justin Fu, John~D. Co{-}Reyes, and Sergey Levine.
\newblock {EX2:} {E}xploration with exemplar models for deep reinforcement
  learning.
\newblock In {\em Proc. of the Conference on Neural Information Processing
  Systems (NeurIPS)}, pages 2577--2587, 2017.

\bibitem[\protect\citeauthoryear{Fu \bgroup \em et al.\egroup
  }{2020}]{fu2020d4rl}
Justin Fu, Aviral Kumar, Ofir Nachum, George Tucker, and Sergey Levine.
\newblock {D4RL:} {D}atasets for deep data-driven reinforcement learning.
\newblock {\em arXiv preprint arXiv:2004.07219}, 2020.

\bibitem[\protect\citeauthoryear{Gregor \bgroup \em et al.\egroup
  }{2017}]{gregor2016variational}
Karol Gregor, Danilo~Jimenez Rezende, and Daan Wierstra.
\newblock Variational intrinsic control.
\newblock In {\em Proc. of the International Conference on Learning
  Representations (ICLR), Workshop Track}, 2017.

\bibitem[\protect\citeauthoryear{Hansen \bgroup \em et al.\egroup
  }{2020}]{hansen2019fast}
Steven Hansen, Will Dabney, Andr{\'{e}} Barreto, David Warde{-}Farley, Tom~Van
  de~Wiele, and Volodymyr Mnih.
\newblock Fast task inference with variational intrinsic successor features.
\newblock In {\em Proc. of the International Conference on Learning
  Representations (ICLR)}, 2020.

\bibitem[\protect\citeauthoryear{Hazan \bgroup \em et al.\egroup
  }{2019}]{hazan2019provably}
Elad Hazan, Sham~M. Kakade, Karan Singh, and Abby~Van Soest.
\newblock Provably efficient maximum entropy exploration.
\newblock In {\em Proc. of the International Conference on Machine Learning
  (ICML)}, pages 2681--2691, 2019.

\bibitem[\protect\citeauthoryear{Laskin \bgroup \em et al.\egroup
  }{2021}]{laskin2021urlb}
Michael Laskin, Denis Yarats, Hao Liu, Kimin Lee, Albert Zhan, Kevin Lu,
  Catherine Cang, Lerrel Pinto, and Pieter Abbeel.
\newblock {URLB:} {U}nsupervised reinforcement learning benchmark.
\newblock In {\em Proc. of the Conference on Neural Information Processing
  Systems (NeurIPS), Datasets and Benchmarks Track}, 2021.

\bibitem[\protect\citeauthoryear{Laskin \bgroup \em et al.\egroup
  }{2022}]{laskin2022cic}
Michael Laskin, Hao Liu, Xue~Bin Peng, Denis Yarats, Aravind Rajeswaran, and
  Pieter Abbeel.
\newblock Unsupervised reinforcement learning with contrastive intrinsic
  control.
\newblock In {\em Proc. of the Conference on Neural Information Processing
  Systems (NeurIPS)}, pages 34478--34491, 2022.

\bibitem[\protect\citeauthoryear{Lee \bgroup \em et al.\egroup
  }{2021}]{lee2021sunrise}
Kimin Lee, Michael Laskin, Aravind Srinivas, and Pieter Abbeel.
\newblock {SUNRISE:} {A} simple unified framework for ensemble learning in deep
  reinforcement learning.
\newblock In {\em Proc. of the International Conference on Machine Learning
  (ICML)}, pages 6131--6141, 2021.

\bibitem[\protect\citeauthoryear{Liu and Abbeel}{2021a}]{liu2021aps}
Hao Liu and Pieter Abbeel.
\newblock {APS:} {A}ctive pretraining with successor features.
\newblock In {\em Proc. of the International Conference on Machine Learning
  (ICML)}, pages 6736--6747, 2021.

\bibitem[\protect\citeauthoryear{Liu and Abbeel}{2021b}]{liu2021behavior}
Hao Liu and Pieter Abbeel.
\newblock Behavior from the void: Unsupervised active pre-training.
\newblock In {\em Proc. of the Conference on Neural Information Processing
  Systems (NeurIPS)}, pages 18459--18473, 2021.

\bibitem[\protect\citeauthoryear{Mutti \bgroup \em et al.\egroup
  }{2021}]{mutti2021task}
Mirco Mutti, Lorenzo Pratissoli, and Marcello Restelli.
\newblock Task-agnostic exploration via policy gradient of a non-parametric
  state entropy estimate.
\newblock In {\em Proc. of the {AAAI} Conference on Artificial Intelligence
  (AAAI)}, pages 9028--9036, 2021.

\bibitem[\protect\citeauthoryear{Oord \bgroup \em et al.\egroup
  }{2018}]{oord2018representation}
Aaron van~den Oord, Yazhe Li, and Oriol Vinyals.
\newblock Representation learning with contrastive predictive coding.
\newblock {\em arXiv preprint arXiv:1807.03748}, 2018.

\bibitem[\protect\citeauthoryear{Park \bgroup \em et al.\egroup
  }{2022}]{park2021lipschitz}
Seohong Park, Jongwook Choi, Jaekyeom Kim, Honglak Lee, and Gunhee Kim.
\newblock Lipschitz-constrained unsupervised skill discovery.
\newblock In {\em Proc. of the International Conference on Learning
  Representations (ICLR)}, 2022.

\bibitem[\protect\citeauthoryear{Park \bgroup \em et al.\egroup
  }{2023}]{park2023controllability}
Seohong Park, Kimin Lee, Youngwoon Lee, and Pieter Abbeel.
\newblock Controllability-aware unsupervised skill discovery.
\newblock In {\em Proc. of the International Conference on Machine Learning
  (ICML)}, pages 27225--27245, 2023.

\bibitem[\protect\citeauthoryear{Pathak \bgroup \em et al.\egroup
  }{2017}]{pathak2017curiosity}
Deepak Pathak, Pulkit Agrawal, Alexei~A. Efros, and Trevor Darrell.
\newblock Curiosity-driven exploration by self-supervised prediction.
\newblock In {\em Proc. of the International Conference on Machine Learning
  (ICML)}, pages 2778--2787, 2017.

\bibitem[\protect\citeauthoryear{Pathak \bgroup \em et al.\egroup
  }{2019}]{pathak2019self}
Deepak Pathak, Dhiraj Gandhi, and Abhinav Gupta.
\newblock Self-supervised exploration via disagreement.
\newblock In {\em Proc. of the International Conference on Machine Learning
  (ICML)}, pages 5062--5071, 2019.

\bibitem[\protect\citeauthoryear{Seo \bgroup \em et al.\egroup
  }{2021}]{seo2021state}
Younggyo Seo, Lili Chen, Jinwoo Shin, Honglak Lee, Pieter Abbeel, and Kimin
  Lee.
\newblock State entropy maximization with random encoders for efficient
  exploration.
\newblock In {\em Proc. of the International Conference on Machine Learning
  (ICML)}, pages 9443--9454, 2021.

\bibitem[\protect\citeauthoryear{Sharma \bgroup \em et al.\egroup
  }{2020}]{sharma2019dynamics}
Archit Sharma, Shixiang Gu, Sergey Levine, Vikash Kumar, and Karol Hausman.
\newblock Dynamics-aware unsupervised discovery of skills.
\newblock In {\em Proc. of the International Conference on Learning
  Representations (ICLR)}, 2020.

\bibitem[\protect\citeauthoryear{Yang \bgroup \em et al.\egroup
  }{2023}]{yang2023behavior}
Rushuai Yang, Chenjia Bai, Hongyi Guo, Siyuan Li, Bin Zhao, Zhen Wang, Peng
  Liu, and Xuelong Li.
\newblock Behavior contrastive learning for unsupervised skill discovery.
\newblock In {\em Proc. of the International Conference on Machine Learning
  (ICML)}, pages 39183--39204, 2023.

\bibitem[\protect\citeauthoryear{Zhang \bgroup \em et al.\egroup
  }{2021}]{zhang2021made}
Tianjun Zhang, Paria Rashidinejad, Jiantao Jiao, Yuandong Tian, Joseph~E.
  Gonzalez, and Stuart Russell.
\newblock {MADE:} {E}xploration via maximizing deviation from explored regions.
\newblock In {\em Proc. of the Conference on Neural Information Processing
  Systems (NeurIPS)}, pages 9663--9680, 2021.

\bibitem[\protect\citeauthoryear{Zhao \bgroup \em et al.\egroup
  }{2022}]{zhao2022mixture}
Andrew Zhao, Matthieu~Gaetan Lin, Yangguang Li, Yong{-}Jin Liu, and Gao Huang.
\newblock A mixture of surprises for unsupervised reinforcement learning.
\newblock In {\em Proc. of the Conference on Neural Information Processing
  Systems (NeurIPS)}, pages 26078--26090, 2022.

\end{thebibliography}

\end{document}